\documentclass[10pt,twocolumn,letterpaper]{article}

\usepackage{iccv}
\usepackage{times}
\usepackage{epsfig}
\usepackage{graphicx}
\usepackage{amsmath}
\usepackage{amssymb}
\usepackage{resizegather}
\usepackage{subcaption}
\usepackage{authblk}

\usepackage[pagebackref=true,breaklinks=true,letterpaper=true,colorlinks,bookmarks=false]{hyperref}

\iccvfinalcopy 

\begin{document}

\title{Split-GCN: Effective Interactive Annotation for Segmentation of Disconnected Instance}

\author[1]{Namgil Kim}
\author[2]{Barom Kang}
\author[2, *]{Yeonok Cho}
\affil[1]{Department of Data Science, Ajou University}
\affil[2]{SelectStar AI Research}
\affil[*]{Corresponding author:   imyeon@selectstar.ai}

\maketitle

\begin{abstract}
   Annotating object boundaries by humans demands high costs. Recently, polygon-based annotation methods with human interaction have shown successful performance. However, given the connected vertex topology, these methods exhibit difficulty predicting the disconnected components in an object. This paper introduces Split-GCN, a novel architecture based on the polygon approach and self-attention mechanism. By offering the direction information, Split-GCN enables the polygon’s vertices to move more precisely to the object boundary. Our model successfully predicts disconnected components of an object by transforming the initial topology using the context exchange about the dependencies of vertices. Split-GCN demonstrates competitive performance with the state-of-the-art models on Cityscapes and even higher performance with the baseline models. On four cross-domain datasets, we confirm our model's generalization ability.  
\end{abstract}
\vspace{-4mm}
\section{Introduction}
Recently, the data-driven deep learning techniques have gained popularity by proving outstanding performance in object segmentation tasks in various fields \cite{bai2017deep,deformablegrid,  hu2017deep,liu2018path, neuhold2017mapillary}. Object segmentation is a problem of classifying objects in pixel units into a given class. However, object segmentation is challenging as networks should distinguish even the object’s contours in the image, which typically contains scale variability, occlusion, and motion blur. To improve such a neural network’s performance on the object segmentation, it is necessary to train with larger amounts and more scale of annotated data. Unfortunately, annotating high-quality ground-truth data for object segmentation is heavily time-consuming. For instance, it takes $40$ seconds for humans to annotate a single object in an image \cite{acuna2018efficient, chen2014beat}. 

There are several previous studies on user interactive annotation methodologies such as scribbles \cite{bai2014error, boykov2001interactive}, clicks \cite{maninis2018deep, papadopoulos2017extreme,  wang2019object, xu2016deep}, and polygons \cite{acuna2018efficient, castrejon2017annotating,  ling2019fast}. Scribbles allow users to draw and drag lines on the region of interest to adjust. On the other hand, clicks request users to click instead of drawing, which is similar to the scribbles but less burdens. However, both methods are different from how users generally annotate. This causes users to learn how to annotate in order to make a detailed correction.

The polygon-based methods are more practical since these methods resemble the user's actual process in the annotation. The concept of these methods is that contours are connected according to the sequence of the points like the Eulerian path. However, because of such connected contours, they have limitations in expressing the disconnected target object occluded by other objects \cite{acuna2018efficient, ling2019fast}. This is a fatal drawback since, in practical usage, users regularly encounter separated components of an object. 


In this paper, we introduce \textit{Split-GCN}, which tackles the problem of interactive annotations for object segmentation. Split-GCN effectively transforms the initial polygon topology into the target object topology, even if it is disconnected. Based on the graph convolutional neural network (GCN), Split-GCN frames the shape of the object by moving every vertex of a polygon considering its position and using the dependencies on each vertex's connected adjacent vertices simultaneously. 

Our model exploits the \textit{motion vector branch} to recognize the direction towards the target object's boundary, similar to looking up a map to find the exact route to an unknown destination. We suggest a novel structure called \textit{separating network} to express the disconnected components split by the occlusion. The separating network not only comprehends the dependencies involved in the context node feature and location of all vertices but also reconstructs the topology of the graph to express the object's shape containing the disconnected components.



\begin{figure*}[ht!]
    \centering
    \includegraphics[width=\textwidth]{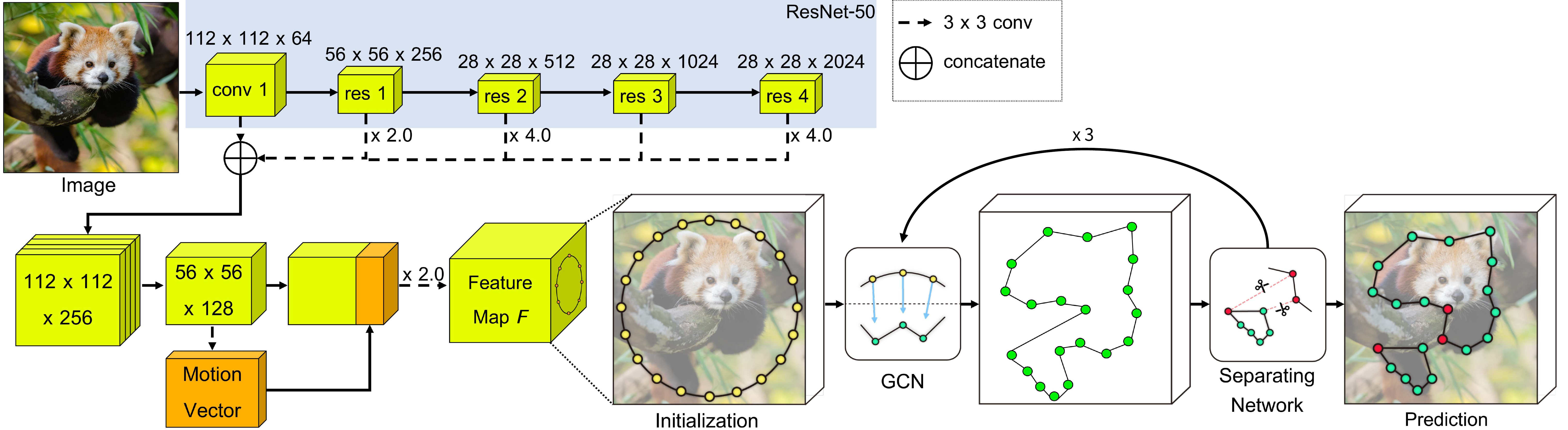}
    \caption{\textbf{Overview of Split-GCN.} Given the feature map $F$ of the cropped image, Split-GCN first extracts the feature at each initialized vertex and moves the vertex's offsets to the target boundary using GCN. Then, the separating network of our model reconstructs graph topology by splitting and merging the vertices. While repeating the process of GCN to Separating network three times,  it yields a more sophisticated ﬁnal predicted vertices to apply separating network output to the adjacency matrix of GCN. Red vertices and multiplication symbols denote disconnected vertices and bilinear interpolation, respectively.}
    \vspace{-4mm}
    \label{fig:intro}
\end{figure*}
We conduct various experiments to demonstrate superiority in the performance and generalization ability of our proposed model on the Cityscape and four datasets from the different domains, KITTI \cite{geiger2012we}, Rooftop \cite{sun2014free}, ADE20K \cite{zhou2017scene}, and Card.MR \cite{suinesiaputra2014collaborative}. Our Split-GCN shows $29.6$ AP in automatic mode experiments, which is very competitive compared to the state-of-the-art models on the Cityscapes test set. Moreover, when compared with the baseline models on the Cityscapes validation dataset, Split-GCN shows the highest performance of $76.6$ mIoU, $52.5$ F$_\mathrm{1px}$, and 67.5 F$_\mathrm{2px}$ scores, improving by $2.9$ mIoU, $4.8$ F$_\mathrm{1px}$, and $3.9$ F$_\mathrm{2px}$ scores from the best performing baseline model, Curve-GCN \cite{ling2019fast}. In cross-domain experiments, trained with only $10\%$ of the total dataset, our model outperforms the baseline models in all datasets as well. In summary, our contributions are three-fold:
\begin{itemize}
    \item We introduce a novel architecture named Split-GCN, which effectively transforms the initial polygon topology into the target object topology. 
  \item  To predict the disconnected components more accurately, we develop the separating network using self-attention.
  \item Split-GCN demonstrates $3.9\%$ and $2.6\%$ bigger mIoU on the Cityscapes and cross-domain datasets, respectively than Curve-GCN, which is the latest model using polygon-based methods as our model. 
 \end{itemize}
\section{Related work}

\noindent\textbf{Pixel-based methods:}
Most interactive annotation models adopt a pixel-wise approach, which predicts an instance based on the pixel unit.
Early studies introduced the methods that scribble instance to discriminate a foreground/background and use graph-cuts to segment \cite{boykov2001experimental,boykov2001interactive}. Later, Rother \textit{et al}. \cite{rother2004grabcut} proposed GrabCut, which uses graph-cuts iteratively and requests bounding boxes to the user to perform segmentation. 


Many studies have recently applied deep neural networks-based models and achieved exceptional performance in the interactive annotation field \cite{ liew2019multiseg, lin2020interactive, majumder2019content}. The first study using deep architecture was deep interactive object selection \cite{xu2016deep}, which converts positive/negative clicks received by the users into the Euclidean distance map to segment objects of interest. Following \cite{xu2016deep}, Mahadevan \textit{et al.} \cite{mahadevan2018iteratively} improved the performance by iterating added user clicks on the errors in the current segmentation. Utilizing extreme clicks \cite{papadopoulos2017extreme} provided by annotators as inputs, DEXTR \cite{maninis2018deep} achieved outstanding results in the interactive annotation. However, as applying the pixel-wise method, it is challenging for most of these models to distinguish between foreground and background pixels, especially when they are similar.
\\[7pt]
\noindent\textbf{Contour-based methods:}
There are two major contour-based methods: the level sets approach and the polygon-based approach.
The level sets \cite{caselles1997geodesic} method predicts object boundaries by continuously taking derivative on the energy function of the curve and data.
Ping Hu \textit{et al.} \cite{hu2017deep} derived the level sets function using a convolutional neural network (CNN). Marcos \textit{et al.}    \cite{marcos2018learning} and Wang \textit{et al.} \cite{wang2019object} optimized CNN based level sets in end-to-end fashion.\\
\indent One of the earliest polygon-based annotation studies was Polygon-RNN \cite{castrejon2017annotating}, which sequentially predicts vertices of an instance one by one using CNN and the recurrent neural network (RNN). To solve the low-resolution accuracy from \cite{castrejon2017annotating}, PolygonRNN++ \cite{acuna2018efficient} improved the performance by applying a gated graph sequence neural network \cite{li2015gated} to increase the output resolution. Curve-GCN \cite{ling2019fast} used the GCN to predict all vertices at once and significantly saved the inference time than PolygonRNN++. However, Curve-GCN shows a limited prediction of disconnected objects because it keeps the connected initial topology until the final prediction. \\
\indent Our proposed model follows the polygon-based approach. We directly predict the object boundary using numerous uniformly spaced points. Especially, our approach allows an efficient prediction for the split of edge instances by transforming the initial topology. 
\begin{figure}[t]
    \centering
    \includegraphics[width=0.45\textwidth]{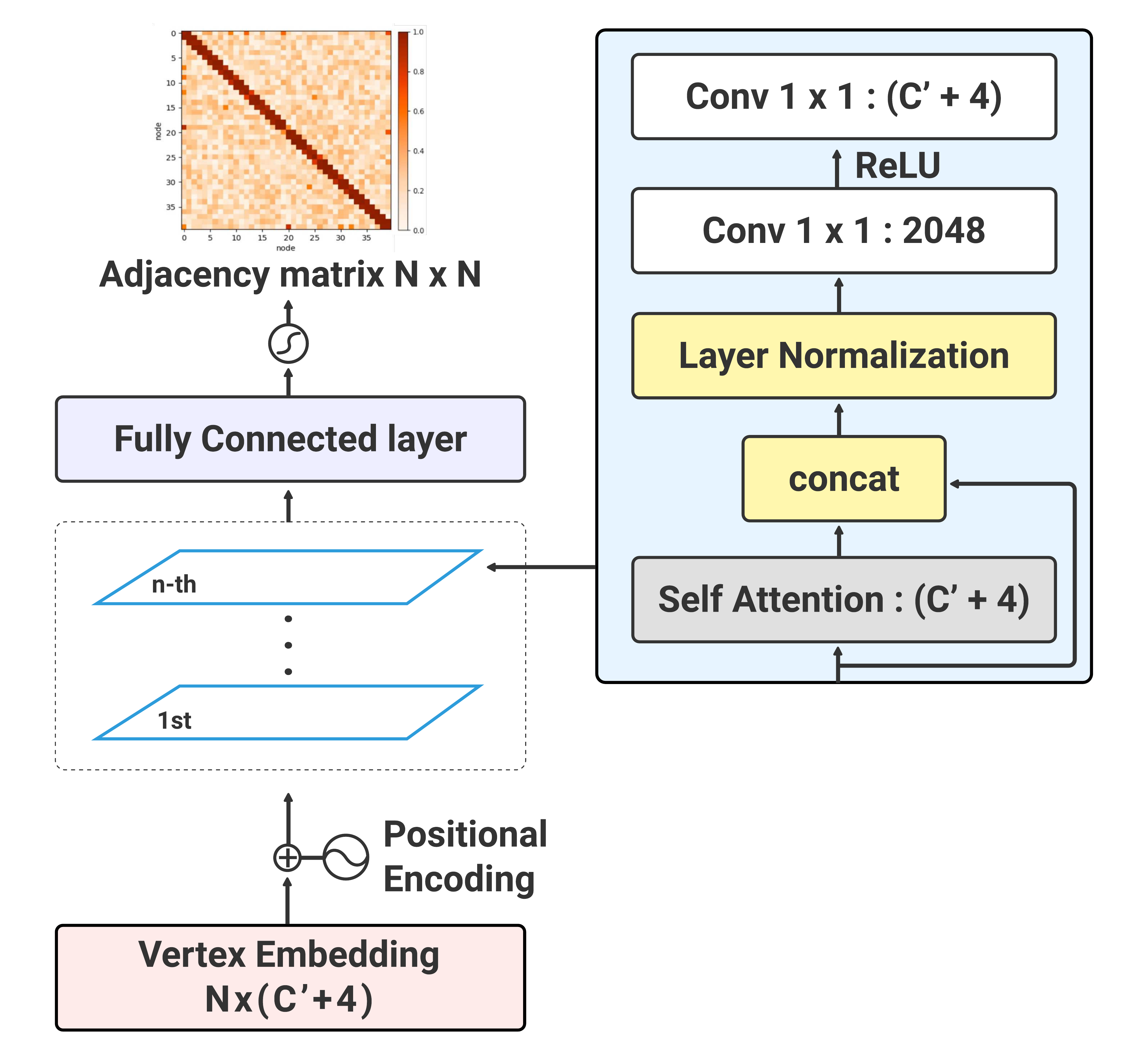}
    \caption{\textbf{Separating network architecture.}  } 
    \vspace{-4mm}
    \label{fig:separating architecture}
\end{figure}
\section{Split-GCN}
Our goal is to develop a model that can express the disconnected components of an object's polygon shape that is occluded by another one using only a few control points. We introduce Split-GCN, a novel deep architecture that consists of the polygon-based approach \cite{acuna2018efficient, ling2019fast} and self-attention mechanisms \cite{vaswani2017attention}. Split-GCN intuitively outlines an object by connecting points in sequential order. We follow a representative labeling scenario from \cite{acuna2018efficient, ling2019fast}. We receive the cropped object as input and predict its shape as a regression problem by shifting offsets. Simultaneously, our model reconstructs the adjacency matrix to predict either connected or disconnected components of the object. 

As shown in Figure \ref{fig:intro}, the Split-GCN primarily consists of two parts: an encoder (feature extraction network) to extract the boundary information of an object and a decoder (novel graph composition network) to capture the shape of an object. We first explain the encoder architecture for feature extraction in Section \ref{subsec:Feature Extraction Network} and then explain the decoder for capturing disconnected points of an object in Section \ref{subsec:Graph Composition Network}. In Section \ref{subsec:Training}, we explain how to train our model. In Section \ref{subsec:Interactive Inference} and, finally, we describe the details of interactive inference, which interacts with a human to perform semi-auto labeling.

\subsection{Feature Extraction Network}\label{subsec:Feature Extraction Network}
The feature extraction network aims to provide efficient object boundary information so that the decoder can express a deforming polygon. We extract high-level semantic information from the object of interest using CNN. We also extract motion vector information, which guides the polygon to the object's boundary at each pixel using the motion vector branch.
\\[7pt]
\noindent\textbf{Feature Extraction:}
For extracting features, we first resize the bounding box generated by the annotator to $224\times 224$ and then encode the cropped bounding box to high-quality semantic information using CNN. Following \cite{acuna2018efficient, ling2019fast}, we deploy ResNet-50 \cite{he2016deep} as a backbone of our feature extraction network. However, we remove the average pooling and fully connected layers to attach our proposed decoder, which captures the object's polygon shape. Then, we adopt the residual encoder architecture from \cite{acuna2018efficient}, which increases the resolution of feature maps without reducing the receptive field. 

As a result, we obtain high-quality semantic information while maintaining low-level details simultaneously. As shown in Figure \ref{fig:intro}, we use bilinear interpolation as the highest possible resolution size at each convolution stage before concatenation. After concatenation, we produce feature maps denoted as $F_{c}\in\mathbb{R}^{C'\times H'\times W'}$ by using $3\times 3$ convolutional filters with stride $2$, batch normalization \cite{ioffe2015batch}, and ReLU activation function where $C'$, $H'$ and $W'$ are indicate channel, height, and width of $F_{c}$ respectively.
\\[7pt]
\noindent\textbf{Motion Vector Branch:}
Our proposed framework moves vertices of the polygon to target boundary in the decoder. We assume that the model can efficiently reach an object's boundary if it has direction information. Motion vector branch exploits one convolutional layer and one fully connected layer to predict the ground truth (GT) motion map $\vec U_\mathrm{gt}\in \mathbb{R}^{2 \times H'\times W'}$, which is the direction to the boundary. In other words, the motion vector branch predicts motion map $\vec V_{\theta}\in \mathbb{R}^{2 \times H'\times Wf'}$ with the magnitude and direction of the object boundary by utilizing $F_{c}$. We denote $\theta$ as the network's parameters. 

We design a GT motion map $\vec U_\mathrm{gt}$ with two channels by differentiating each coordinate of $(x, y)$ respectively. $\vec U_\mathrm{gt}$ is 
\begin{equation} 
\vec U_\mathrm{gt}(x,y)=-\frac{\nabla \phi_\mathrm{DT}(x, y) }{|\nabla \phi_\mathrm{DT}(x, y)|},
\end{equation}
where $\phi_{\mathrm{DT}}(x,y) = \min_{(x_{\mathrm{gt}}, y_{\mathrm{gt}})\in G}\|(x,y) - (x_{\mathrm{gt}}, y_{\mathrm{gt}})\|_{2}$ for $(x_{\mathrm{gt}}, y_{\mathrm{gt}})\in G $ and $G$ is the set of GT boundary points. In practice, the model has difficulty predicting the GT boundary corresponding to the object because the boundary has the size of only one pixel. Thus, we use the 9$\times$9 Gaussian filter to make the boundary thicker.  



\subsection{Graph Composition Network }\label{subsec:Graph Composition Network}
The graph composition network intuitively renders the target boundary using $N$ control points and identifies disconnected control points to express the split component. Here, intuitive rendering means connecting points one by one in a straight line. Similar to \cite{ling2019fast, wang2018pixel2mesh}, we define each control point as a random variable and use the GCN to find the correct position of the control point through the relation of nearby points that contains the cues of the target object. We grasp the dependencies among the vertices using our proposed separating network; we reconstruct graph topology and predict disconnected points following these dependencies. We utilize a self-attention Transformer network \cite{vaswani2017attention} to take advantage of such complicated dependencies. 
\\[7pt]
\noindent\textbf{Vertex Embedding:}
The vertex embedding produces features for the vertex in the GCN. First, we normalize each channel of the predicted motion map $\vec V_{\theta}$ by $H'$ and $W'$. Then, we concatenate $F_{c}$ with these two channels to guide the network to head toward the target boundary. Moreover, as low resolution produces blocky polygons, especially for large objects, we resize the concatenated feature to 112$\times$112, exploiting the bilinear interpolation. We denote this embedding as $F$ (see Figure \ref{fig:intro}).\\[7pt]
\noindent\textbf{GCN:} The graph of GCN model is denoted as $G = (V, E)$ where the vertices are $V$, and the edges are $E$. Here, we construct $V$ and $E$ to represent control points and adjacency matrix, respectively. This adjacency matrix is inferred from the separating network, and the GCN model exploits it to exchange and propagate information between nodes in the graph.\vphantom{The graph of GCN model consists of $V$ and $E$ denoted as $G = (V, E)$ signifying the vertices as $V$ and the edges as $E$ in the graph. We construct $V$ and $E$ to represent control points and adjacency matrix, respectively. The GCN model exploits the adjacency matrix, a result inferred from the separating network, to exchange and propagate information between nodes in the graph.}
The GCN model shifts $N$ control points to be located uniformly at the object's boundary through feature exchange across the $N$ feature vectors. The feature vectors are the coordinate information of $N$ control points extracted from vertex embedding. Here, the initial positions of these $N$ control points are formed to be a fixed circular topology, which covers around $75\%$ of the target image.

We denote the predicted points set as $\boldsymbol{P} = \left\{p_{i} = (x_{i}, y_{i}) \right\}_{i=0,1, \cdots, N-1}$ and the GT points set as $\boldsymbol{Q} = \left\{q_{i} = (x'_{i}, y'_{i}) \right\}_{i=0,1, \cdots, N-1}$, where $N$ is the number of points. The GT points set is uniformly extracted from the boundary of an object.\vphantom{The GT control points (GT points) set is uniformly extracted from the boundary of an object. We denote the predicted control points (predicted points) set as $\boldsymbol{P} = \left\{p_{i} = (x_{i}, y_{i}) \right\}_{i=0,1, \cdots, N-1}$ and the GT points set as $\boldsymbol{Q} = \left\{q_{i} = (x'_{i}, y'_{i}) \right\}_{i=0,1, \cdots, N-1}$, where $N$ is the number of points.} Following \cite{ling2019fast, wang2018pixel2mesh}, we use Graph-ResNet to enable a more precise prediction by exploiting shortcut connections.
The feature vector is defined as
\begin{equation}\label{gcn}
 f^{l}_{\left(x_{i},y_{i}\right)} = \mathrm{concat}\left\{F(x_{i}, y_{i}), x_{i}, y_{i}, \Delta_{H} x_{i}, \Delta_{H} y_{i}\right\},
\end{equation}
where $i$ and $l$ are the current index and layer respectively. We let  $x_{i}, y_{i} \in [0,1]$ be the normalized current points. $\Delta_{H}$ denotes human interactive inference, where $\Delta_{H} x_{i}$ and $\Delta_{H} y_{i}$ represent the distance from the ground truth. If there is no interactive inference, then $\Delta_{H} x_{i}$ and $\Delta_{H} y_{i}$ are set to be zeros.
With the defined feature vector $f^{l}_{(x_{i},y_{i})}$, the internal mechanism of the Graph-ResNet follows the steps below:
\begin{equation}
r^{l}_{p_{{i}}}=\sigma\left(w^l_{0}f^l_{p_{i}}+\sum_{z\in \boldsymbol{N}(p_{i})}w^l_{1}f_z^l\right),
\end{equation}
\begin{equation} 
r^{l+1}_{p_{{i}}}= \tilde{w}^{l}_{0}r^l_{p_{{i}}}+\sum_{z \in \boldsymbol{N}(p_{i})} \tilde{w}^{l}_{1}r_z^l.
\end{equation}
From $r^{l+1}_{p_{{i}}}$ and $f^l_{p_{{i}}}$, we have the next feature vector
\begin{equation}
    f^{l+1}_{p_{{i}}} = \sigma\left(r^{l+1}_{{p_{{i}}}} + f^l_{p_{{i}}}\right),
\end{equation}
where $w^{l}$ and $\tilde{w}^{l}$ are the weight matrices. 
 

 We adopt a fully connected layer on the last layer of the GCN to obtain $\Delta x_{i}$ and $\Delta y_{i}$ that are shifts needed to move from its current location. Then, the next GCN layer re-extracts the feature vectors from newly shifted location, $[x_{i}+\Delta x_{i},y_{i}+\Delta y_{i}]$.
\\[7pt]
\noindent\textbf{Separating Network:}
In SplitGCN, the separating network, which consists of self-attention of the Transformer and a fully connected layer, decomposes the graph into connected components. More specifically, the separating network takes a feature vector $f\in \mathbb{R}^{N\times(C'+2+2)}$ extracted from the position coordinates predicted by GCN (see Figure \ref{fig:intro}) as an input and predicts an adjacency matrix as an output. In particular, a feature vector is the concatenation of feature map $F$  and the coordinate vector ($\Delta_{H}x,\Delta_{H}y$) from Human interactive inference. As mentioned earlier, the feature map $F$ consists of two vectors: the feature map $F_{c}$ and the motion vector. The feature map $F_{c}$  provides semantic perspective information on node similarity. The motion vector provides directional information at particular boundaries. If two points belong to separated components, the direction of those corresponding motion vectors will be opposite  to each other, despite their proximity.

We employ three feed-forward networks to obtain matrices $Q(f)$, $K(f)$, and $V(f)$. We first multiply $Q(f)$ by $K(f)$, and to avoid gradients too small, we divide the result by scale factor $d_{k}$, which has the dimension of $K(f)$ (the same dimension as $Q(f)$). Finally, we take softmax over it which is then multiplied by $V(f)$ to obtain the dependencies between all points and point-wise. We can write such self-attention mechanism as
\begin{equation}
\mathrm{Attn}(Q(f),\!K(f),\!V(f))\!=\! \mathrm{softmax}\!\left(\frac{Q(f)K^{\top}(f)}{\sqrt{d_{k}}}\right)\!V(f).
\end{equation}
The process after the self-attention mechanism follows the encoder in the vanilla Transformer. Since the input of separating network is fixed, we apply the absolute positional encoding with the fixed sinusoidal embedding \cite{vaswani2017attention} to the input of $Q$, $K$ and $V$. As seen in Figure \ref{fig:separating architecture}, we iterate this process six times, which makes the connectedness (or disconnectedness) of neighboring points become more evident. After the last Transformer layer, we use another fully connected layer with a sigmoid function to predict the $N \times N$ adjacency matrix of the vertices. Figure \ref{fig:separating architecture} portrays the architecture of the separating network. As shown in Figure \ref{fig:intro}, GCN can make more accurate predictions using separating network. 
\subsection{Training }\label{subsec:Training}
We train our separating network in an end-to-end manner using the following process. Once trained, the model can recognize the separating point and can draw completely disconnected boundaries of separate components.\\[7pt]
\noindent\textbf{GT Construction:} In this part, we describe how to design a GT adjacency matrix and GT points. First, we define that a split factor $k$ represents the maximum number of disconnected components in an object. For instance, $k = 2$ means that an object consists of one component or two disconnected components. We set the range of $k$ to $1 \leq k$, where $k\in\mathbb{N}$. For the given split factor $k\ge 2$, we assign the $t$--th component to $m=\lfloor{N/k}\rfloor$ points if $1\le t\le k-1$ and the $k$--th component to $m'= N - m(k-1)$. Then, we number points at the $t$--th component in such a way that a simple closed path $m$--gon is clockwise oriented if $t$ is odd and counterclockwise if otherwise. In the same manner, $m'$--gon is clockwise if $k$ is odd and counterclockwise if otherwise.\vphantom{We have a clockwise $m'$--gon if $k$ is odd, counterclockwise if otherwise.} The adjacency matrix of these polygons is a block diagonal matrix with $k$ blocks. Blocks up to $(k-1)$--th have a size $m\times m$; the entries in each block are assigned with one corresponding to edges of $m$--gon.\vphantom{Then, we number points at the $t$--th component in such a way that the $m$--gon ($m'$--gon, a simple closed path) is clockwise oriented if $t$ is odd and counterclockwise if otherwise.}\vphantom{ When $k=1$, GT is clockwise $N$gon.}\vphantom{The first $k-1$ blocks have size $m\times m$ and the entries in each block assigned to 1 correspond to edges of $m$--gon.} The $k$--th block is $m'\times m'$ matrix whose entries follow the rules above. When $k=1$, a single clockwise $N$--gon and its adjacency matrix are obtained. \vphantom{from GT points. In the case of $k=1$, the entries of this matrix represent a single $N$--gon.}\\[7pt]
\noindent\textbf{Polygon Separating Loss:}
%
%
Polygon separating loss finds the disconnected points using the adjacency matrix. An entry of our GT adjacency matrix has a value of $0$ or $1$. If two points are connected to each other, it takes the value $1$, otherwise $0$.  However, as most of these values are $0$, it causes an imbalance problem when training the model. We solve this imbalance by giving a difference ratio $\beta$ of connected and disconnected points in training. We denote an entry of the predicted adjacency matrix by $A_\mathrm{pred}$ and of the GT adjacency matrix by $A_\mathrm{gt}\in [0, 1]$. Polygon separating loss is defined as
\begin{equation}
  L_\mathrm{sep}({A}_\mathrm{pred},{A}_\mathrm{gt}) \!=\!
  \begin{cases}
    -\beta\log({A}_\mathrm{pred}) & \!\!\text{if $A_\mathrm{gt} \!= \!1$} \\
     -(1-\beta)\log(1-{A}_\mathrm{pred}) &\!\! \text{otherwise.}
  \end{cases}
\end{equation}

To acquire the final predicted output, we truncate all other edges except two most probable edges, i.e., each row of the adjacency matrix must have exactly three non-zero entries, including diagonal entries.\\[2pt]

\noindent\textbf{Point Matching Loss:}
 We use point matching loss \cite{ling2019fast} to locate the predicted points around the GT boundary. However, the resulting of the predicted points are ordered in a predetermined direction that may differ from the direction of GT points.
The point matching loss is defined as follows:
\begin{equation}
L_\mathrm{pmatch}(\boldsymbol{P},\boldsymbol{Q}):= \min_{j \in [0 \cdot\cdot\cdot, N-1]}  \sum_{i=0}^{N-1} \left\| p_{i} - q_{\left(j+i\right){\mathrm{mod}N}}\right\|_{1}.
\end{equation}

\noindent\textbf{Point $\boldsymbol{L_{2}}$ Loss:}
To detect the separation in a disconnected object, it is not sufficient to just place the predicted points near the GT boundary. We need to know precisely which of an individual predicted point is close to an individual disconnected point. This can be easily done once we find a fitting match between GT and predicted points. To achieve this, we propose the following point $L_{2}$ loss function.\vphantom{If the object is disconnected, to detect the separation, just having the predicted boundary closed to GT boundary is not sufficient. To detect the separation in a disconnected object, it is not sufficient to just place the predicted boundary near the GT boundary. \vphantom{More specifically, we need to know which of the predicted points are close to the separating points.} This can be easily done once we find a fitting match between GT and predicted points, since we can identify GT points around the separating points. To achieve this, we propose the following loss function referred to as point $L_{2}$  loss.}
\begin{equation}
L_{2}(\boldsymbol{P},\boldsymbol{Q}):= \min_{\boldsymbol{P}}  \sum_{i=0}^{N-1} \left\| p_{i} - q_{i}\right\|_{2}.
\end{equation}

Consequently, the usage of point $L_{2}$ loss makes the distance between $p_{i}$ and $q_{i}$ as small as possible for each $i$.\vphantom{In short, the usage of point $L_{2}$  loss makes the distance between $p_{i}$ and $q_{i}$ as small as possible for each $i$. While Brute-force search is a possible method, we show competitive performance in our experiments using point $L_{2}$  loss.\vphantom{While it is possible to find the matching using Brute-force search, using point $L_{2}$  loss, we show competitive performance in our experiments.} }
\\[7pt]
\noindent\textbf{Motion Vector Loss:}  Here, we consider the motion vector loss as in \cite{bai2017deep, hu2017deep}. The motion vector loss can be simply regarded as a mean square angular. This angular nature of the motion vector loss enables the model to accurately predict points and be located on the object's exact boundary. Motion vector loss is defined as
\begin{equation}
L_\mathrm{motion}(\theta):=\!\!\!\!{ \sum_{\substack{1\le i\le H',\\ 1\le j\le W'}} \left( \cos^{-1}\left< \frac{\vec V_{\theta}(i, j) }{|\vec V_{\theta}(i, j)|}, {\vec U_\mathrm{gt}(i,j)} \right>\right)^{2}}.
\end{equation}
\vphantom{We train our separating network in an end-to-end manner using the above process. Once trained, the model can recognize the separating point and can draw completely disconnected boundaries of separate components.}
\vphantom{After training separating network including all the above especially, GT structure(directions of GT) and point $L_{2}$  loss(forcing the point ), the model can recognize the separating point, so that it can completely draw two disjoint boundaries of separate components. Indeed, we first train model only with point matching loss and then fine-tune the model with both point $L_{2}$  loss and point matching loss. We experimentally had better performance.}\vphantom{\noindent\textbf{Learning separation:}
Since GT Adj. is always block diagonal, predicted adjacency matrix tends to have the GT Adj. form. By our construction of GT Adj., the disconnected points are inferred in each disconnected components. Moreover, each components of GT points having different direction helps detecting separating points. For example, when split factor is $2$, if we traverse through GT points following their order (one is clockwise and the other one is counterclockwise), the resulting graph has a single intersection of two edges, so that the graph resembles the shape of $8$. This shape like M\"{o}bius strip might affect to find the separating point.  If the predicted graph has not yet detected the separating point, the same phenomenon does not occur in the predicted model. In this case, after applying separating network followed by our structure, the model can recognize the separating point, so that it can completely draw two disconnected boundaries of separate components.} 
\vspace{-5mm}
\subsection{Interactive Inference}\label{subsec:Interactive Inference}
The goal of interactive inference is for a model to generate refined performance through training, similar to the sophisticated modification generated by humans. When a person modifies the vertices predicted by the model, we assume that the most efficient approach is selecting the furthest vertex of the prediction and relocating it to the GT coordinate. 

To implement this, we first calculate the difference of the most distant vertex in the predicted points set \textbf{\textit{P}} and GT points set \textbf{\textit{Q}} using the Manhattan distance. Then, we substitute $\Delta_{H} x_{i}$ and $\Delta_{H} y_{i}$ of the feature vector with the calculated difference. Consequently, by propagating the calculated difference information to connected adjacent vertices, our model adjusts the coordinates of the connected adjacent vertices.
\begin{table*}[t]
\begin{center}
\resizebox{15cm}{!}{
\begin{tabular}{|l|c|c|cccccccc|}
\hline
Model &Train dataset &AP &Person&Rider&Car&Truck&Bus&Train&Motorcycle&Bicycle \\
\hline\hline 
PANet \cite{liu2018path}&\texttt{fine+COCO}
&36.4&41.5&33.6&58.2&31.8&45.3&28.7&28.2&24.1\\ 
Axial-DL-L \cite{wang2020axial} &\texttt{fine+MV} &38.1&34.7&30.4&55.1&40.9&49.7&\textbf{43.5}&29.0&21.7\\
Pan-DL \cite{cheng2020panoptic}&\texttt{fine+MV} &39.0&36.0&30.2&56.7&\textbf{41.5}&\textbf{50.8}&42.5&30.4&23.7\\
LevelSet R-CNN \cite{homayounfar2020levelset}&\texttt{fine+COCO} &40.0&\textbf{43.4}&33.9&\textbf{59.0}&37.6&49.4&39.4&\textbf{32.5}&\textbf{24.9}\\
PolyTransform \cite{liang2020polytransform}&\texttt{fine+COCO} &\textbf{40.1}&42.4&\textbf{34.8}&58.5&39.8&49.9&41.3&30.9&23.4\\
\hline\hline
PolygonRNN++ \cite{acuna2018efficient}&\texttt{fine}&25.5&29.4&21.8&48.3&21.1&32.3&23.7&13.6&13.6\\
Mask R-CNN \cite{he2017mask}&\texttt{fine} &26.2&30.5&23.7&46.9&22.8&32.2&18.6&19.1&16.0\\
BShapeNet+ \cite{kang2020bshapenet}&\texttt{fine} &27.3&29.7&23.4&46.7&26.1&33.3&24.8&20.3&14.1\\
GMIS \cite{liu2018affinity}&\texttt{fine} &27.3&31.5&25.2&42.3&21.8&37.2&28.9&18.8&12.8\\
Neven et al. \cite{neven2019instance}&\texttt{fine} &27.6&34.5&26.1&52.4&21.7&40.9&30.9&24.1&18.7\\
AdaptIS \cite{sofiiuk2019adaptis}&\texttt{fine} &32.5&31.4&\textbf{29.1}&50.0&31.6&41.7&\textbf{39.4}&\textbf{24.7}&12.1\\
SSAP \cite{gao2019ssap}&\texttt{fine} &\textbf{32.7}&\textbf{35.4}&25.5&\textbf{55.9}&\textbf{33.2}&\textbf{43.9}&31.9&19.5&16.2\\
\hline
Split-GCN($k$=1) &\texttt{fine} &29.1&32.7&25.7&49.5&28.0&32.7&21.2&22.2&17.3\\
Split-GCN($k$=3) &\texttt{fine} &29.6&33.2&27.7&48.1&26.6&33.7&25.8&22.9&\textbf{19.3}\\
\hline 
\end{tabular}}
\end{center}
\vspace{-5mm}
\caption{\label{tab_sota}\textbf{Instance segmentation results} on the Cityscapes test set. Comparison of our model to the state-of-the-art models trained with the Cityscapes (\texttt{fine}) or other datasets, MS COCO (\texttt{COCO}) \cite{lin2014microsoft} and Mapillary Vitas (\texttt{MV}) \cite{neuhold2017mapillary}.} 
\vspace{-2mm} 
\end{table*}
\vspace{2mm}
\begin{table*}
\centering
\makebox[0pt][c]{\parbox{\textwidth}{%
				\begin{minipage}[tb]{0.2\hsize}\centering
					\begin{tabular}{|l|c|c|c|}
						\hline
						Model &mIoU&F$_\mathrm{1px}$ &F$_\mathrm{2px}$  \\
						\hline\hline
						PolygonRNN++  &71.4&46.6&62.3\\
						PSP-DeepLab&73.7&47.1&62.8\\
						Spline-GCN &73.7&47.7&63.6\\
						\hline
						Split-GCN($k=1$)&75.3&51.2&66.1\\ 
						Split-GCN($k=3$)&\textbf{76.6}&\textbf{52.5}&\textbf{67.5}\\ 
						\hline 
					\end{tabular}
					\captionsetup{width=1.75\linewidth}
					\vspace{-2mm}
					\caption{\textbf{Comparison of our model to the baseline models} on the Cityscapes validation set.  }
					\label{tab_auto}
				\end{minipage}
				\hfill
				\begin{minipage}[tb]{0.32\hsize}\centering
					\begin{tabular}{|l|c|c|c|}
						\hline
						Split &mIoU&F$_\mathrm{1px}$ &F$_\mathrm{2px}$  \\
						\hline\hline
						$k=1$ &75.3&51.2&66.1\\
						$k=2$ &76.1&51.9&66.8\\
						$k=3$ &\textbf{76.6}&\textbf{52.5}&\textbf{67.5}\\
						$k=4$ &76.2&52.1&67.1\\
						$k=5$ &75.9&52.0&66.5\\
						\hline 
					\end{tabular}
					\captionsetup{width=.9\linewidth}\vspace{-2mm}
					\caption{\textbf{Performance by split factor $\boldsymbol{k}$} on the Cityscapes validation set.}
					\label{tab_split}
				\end{minipage}
				\begin{minipage}[tb]{0.3\hsize}\centering
					\begin{tabular}{|l|c|c|c|}
						\hline
						Points &mIoU&F$_\mathrm{1px}$ &F$_\mathrm{2px}$  \\
						\hline\hline
						20 pts &71.3&46.4&61.9\\
						30 pts &74.8&50.8&64.7\\
						40 pts &\textbf{76.6}&\textbf{52.5}&\textbf{67.5}\\
						50 pts &76.3&52.2&67.4\\
						60 pts &76.2&52.2&67.1\\
						\hline 
					\end{tabular}
					\captionsetup{width=1\linewidth}
					\vspace{-2mm}
					\caption{\textbf{Performance by the number of control points} on the Cityscapes validation set.}
					\label{tab_vertex}
				\end{minipage}%
		}}  
		\vspace{-2mm}
\end{table*}
\vspace{-2mm}
\begin{table}[!]
\begin{center}
\resizebox{7cm}{!}{
\begin{tabular}{|l|c|c|c|}
\hline
Model &mIoU&F$_\mathrm{1px}$ &F$_\mathrm{2px}$  \\
\hline\hline
GCN&70.8&46.4&61.6\\
+ motion branch&73.1&49.9&65.0\\
+ separating network&76.6&52.5&67.5\\ 
+ interactive learning &\textbf{78.3}&\textbf{53.8}&\textbf{68.7}\\
\hline 
\end{tabular}}
\end{center}
\vspace{-5mm}
\caption{\textbf{Ablation study} of the three modules on the Cityscapes validation set.}
\label{tab_abl}
\vspace{-5mm}
\end{table}
\section{Experimental Results }
We evaluate our model in two aspects: \hspace{-1mm}in-domain and cross-domain annotation settings. For in-domain experiments, we use the Cityscapes dataset \cite{cordts2016cityscapes} as the main benchmark to compare our model's performance. For cross-domain experiments, we assess the generalization ability of our trained model (with the Cityscapes dataset) on several different cross-domain datasets: KITTI \cite{geiger2012we}, Rooftop \cite{sun2014free}, ADE20K \cite{zhou2017scene}, and Card.MR \cite{suinesiaputra2014collaborative}. As assumed in \cite{acuna2018efficient,castrejon2017annotating,ling2019fast}, we suppose that the users provide the ground-truth boxes and use them as the input to the model.  
\\[7pt]
\noindent\textbf{Image Encoder:}
\hspace{-2mm}Following the method from PolygonRNN++ \cite{acuna2018efficient}, we use the ResNet-50 \cite{he2016deep} as the backbone of the image encoder.
\\[7pt]
\noindent \textbf{Training Details:}
We train the model with the structure shown in Figure \ref{fig:intro}. We use ImageNet \cite{deng2009imagenet} pretrained weights only for the image encoder in the feature extraction network. For training, we use point matching loss, point $L_{2}$  loss, motion vector loss, and polygon separating loss in an end-to-end manner.\vphantom{A split factor $k$ represents the maximum number of disconnected components in an object. For instance, $k = 2$ means that an object contains one or two disconnected components. We set the range of $k$ to $1 \leq k$, where $k$ is a natural number.} We set the initial adjacency matrix of GCN to a single block $N$--gon as mentioned in Section \ref{subsec:Training}. \vphantom{Control points $N$ represents the number of the predicted points. When $N = 40$, an object boundary consists of 40 control points.} Experiments regarding the value of $k$ and $N$ are in Section \ref{points&split}. For the most experiments, we set the split factor $k = 3$ and $N = 40$.
\begin{table}[!t]
\begin{center}
\resizebox{6cm}{!}{
\begin{tabular}{|l|c|c|}
\hline
Model &mIoU&Clicks  \\
\hline\hline
Spline-GCN-MBOX \cite{ling2019fast}&77.3&2.4\\
+ one click&80.2&3.6\\
\hline
Split-GCN&78.3&2.0\\ 
+ one click &79.4&3.0\\
+ two clicks &\textbf{81.2}&4.0\\
\hline 
\end{tabular}}
\end{center}
\vspace{-4mm}
\caption{\textbf{Human Interaction}. \vphantom{This table shows the average number of click results which is requested to modify the worst boundary predicted point by the annotator. We compare \textit{MBOX} \cite{ling2019fast} which provides the box around the split object for each component with our model on average clicks.}Comparison of the average number of clicks between our model and \textit{MBOX} \cite{ling2019fast}. }
\label{tab_human}
\vspace{-5mm}
\end{table}

We train our model on a single NVIDIA Titan XP GPU for $150$ epochs. We use the batch size of 24 and the initial learning rate and weight decay of 1e-4. Moreover, the decay on the learning rate is $1/10$ at every $50$ epochs. During training, the input image is cropped using a GT bounding box and appropriately resized with a resolution of $224\times 224$. \vphantom{Additionally,}We use standard augmentation techniques (random rotation, horizontal flip, and scaling) on the data.
\begin{figure*}[!]
    \centering
    \includegraphics[width=400pt]{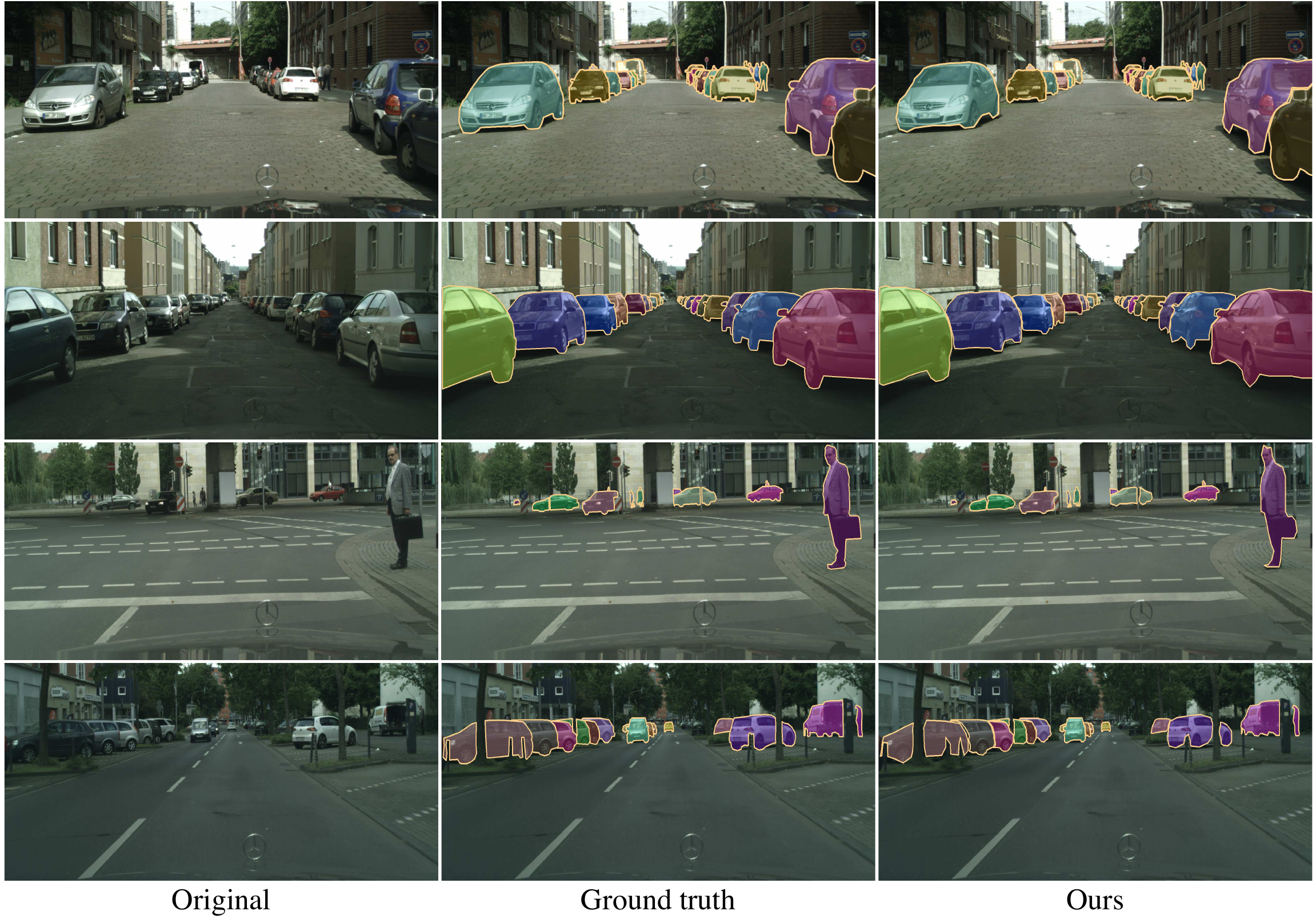}
    \vspace{-4mm}
    \caption{\textbf{Results of Automatic mode on the Cityscapes validation set.} From left to right, original images, ground truth images, and our model predictions on automatic mode.} 
    \label{fig:auto_pic_1}
\end{figure*}
\begin{figure}[!]
    \centering
    \includegraphics[width=\linewidth]{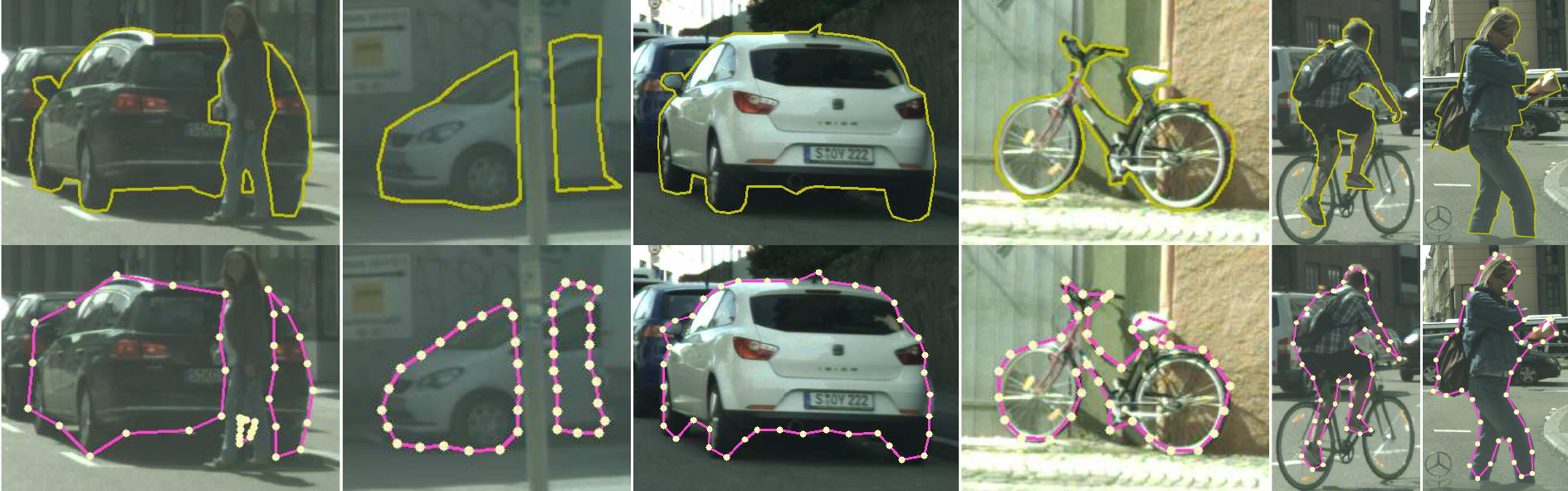}
    \caption{\textbf{Effect of Split factor} $\boldsymbol{k}$. Our model is trained on the Cityscapes validation set with $k=3$. {First and second column show the predictions of three and two disconnected components respectively. The others show the predictions of not occluded single object.} Top: Ground truth and Bottom: Our model results. \vphantom{predicted}  }
    \label{fig:auto_pic_2}
    \vspace{-4mm}
\end{figure}
\\[7pt]
\noindent\textbf{State-of-the-art \& Baseline Models:}
We first compare our model to state-of-the-art models from the Cityscapes leaderboard to demonstrate the competitiveness of our model. We also compare our model to the baseline models. The baseline models are from competitive models using the similar approaches: Curve-GCN (Spline-GCN and Polygon-GCN) \cite{ling2019fast}, PolygonRNN++ \cite{acuna2018efficient}, and PSP-DeepLab \cite{chen2014semantic}. We choose PSP-DeepLab using spatial pyramid pooling \cite{zhao2017pyramid} because it has similar encoder architecture as our model. We also choose Curve-GCN and PolygonRNN++, which are the latest models using polygon-based methodology as our model.
\\[7pt]
\noindent\textbf{Metrics:} 
We use three metrics to evaluate our model: 
(1) We use two Average Precision (AP) metrics to compare the performance of our model to the state-of-the-art models. First, AP is calculated by Intersection-over-Union(IoU) with an increase of $0.05$ from $0.5$ to $0.95$ overlap thresholds. Second, AP$_{50}$ is calculated by IoU with a $0.5$ overlap threshold, which follows the Cityscapes \cite{cordts2016cityscapes} metrics.
(2) We use the mean IoU (mIoU) metric to compare the performance of our model to the baseline models. For each predicted instances and ground truth, we calculated the mIoU by averaging the IoU of every class. (3) As mIoU alone is insufficient to measure the inaccuracies on the object boundary, we use the boundary F score \cite{perazzi2016benchmark} as another metric to evaluate the performance on the object boundary more precisely. The boundary F score measures the performance of the precision and recall on a given boundary width of 1 pixel and 2 pixels, denoted as F$_\mathrm{1px}$ and F$_\mathrm{2px}$, respectively. 

\begin{figure*}[t]
  \centering
    \begin{minipage}{0.48\linewidth}
    \vspace{-2mm}
      \includegraphics[width=1\linewidth]{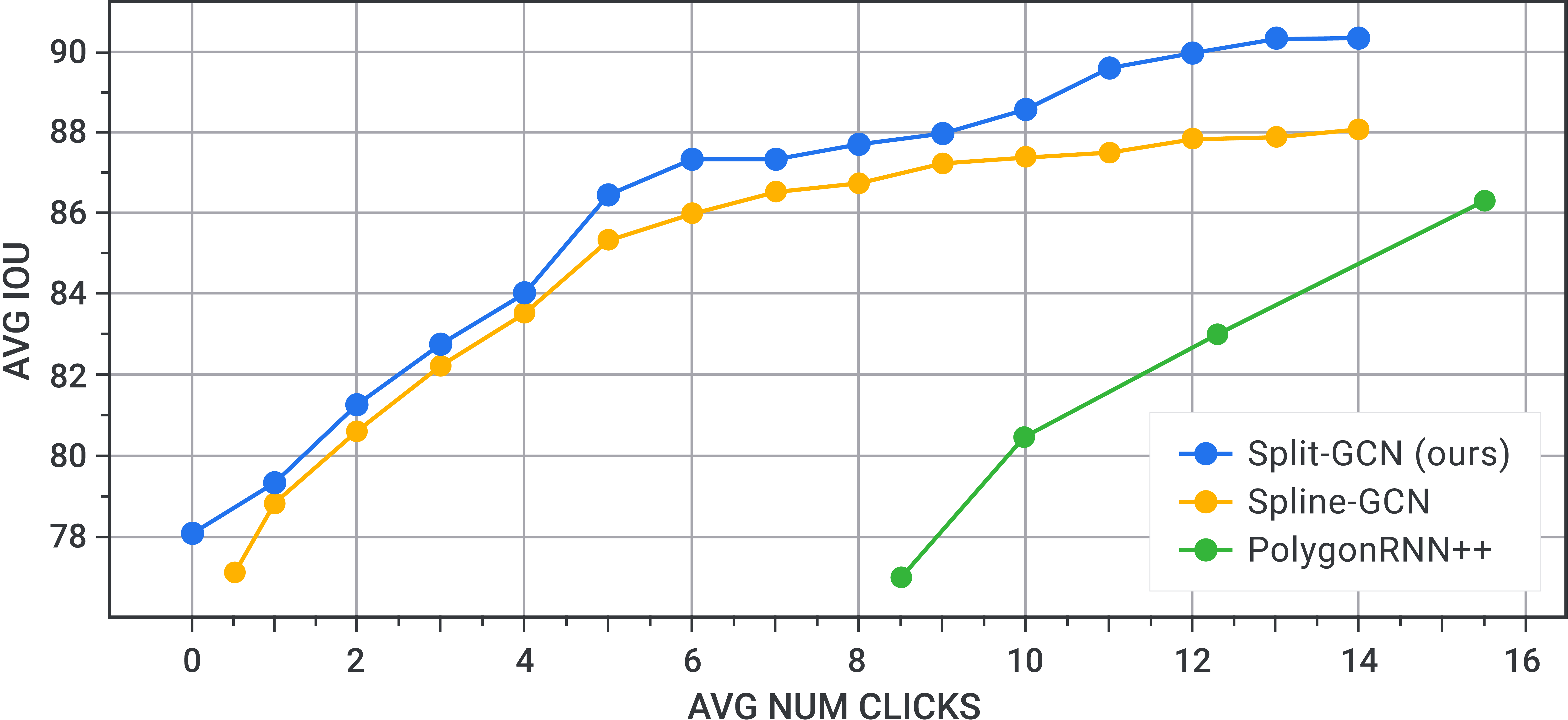}
      \vspace{-7mm}
      \caption{\textbf{Results of Interactive mode} on the Cityscapes validation set.   } 
      \label{fig:graph_1}
    \end{minipage}\hfill
    \begin{minipage}{0.48\linewidth}
      \includegraphics[width=1\linewidth]{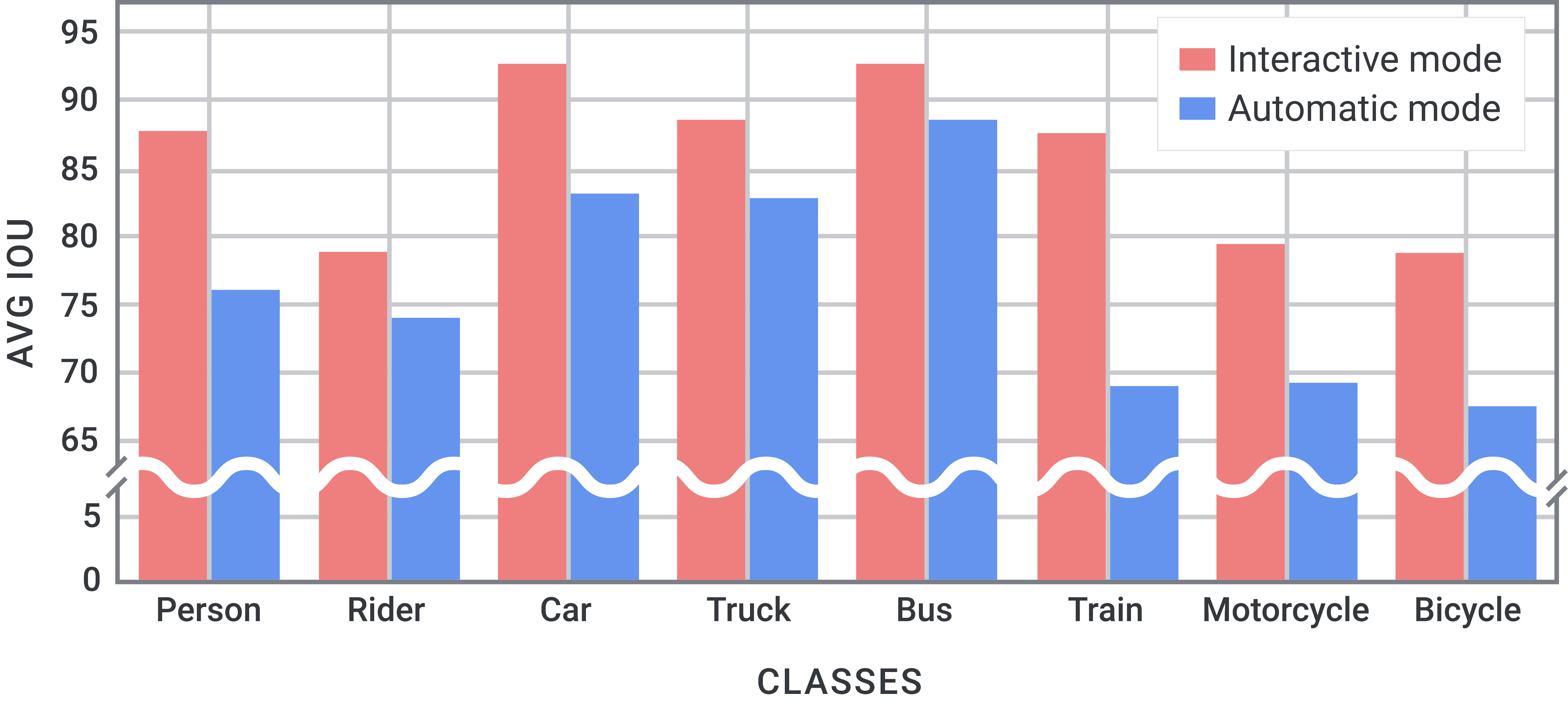}
      \vspace{-7mm}
      \caption{\textbf{Results of Interactive mode and Automatic mode} on the Cityscapes validation set by classes. } 
      \label{fig:graph_2}
    \end{minipage}
    \vspace{-3mm}
\end{figure*}
\begin{table}[!]
\begin{center}
\resizebox{7cm}{!}{
\begin{tabular}{|l|c|c|c|c|}
\hline
Model&KITTI&ADE &Rooftop &Card.MR  \\
\hline\hline
PolygonRNN++&83.1&71.8&65.7&80.6\\
PSP-DeepLab&83.4&72.7&57.9&74.1\\
Polygon-GCN &84.7&77.4&75.6&90.9\\
Spline-GCN &84.8&77.4&78.2&91.3\\
\hline
Split-GCN($k$=1)&85.4&78.1&80.1&92.0\\
Split-GCN($k$=3)&\textbf{86.3}&\textbf{79.2}&\textbf{80.2}&\textbf{92.3}\\
\hline 
\end{tabular}}
\end{center}
\vspace{-3mm} 
\caption{\label{tab_cross}\textbf{Results of cross-domain datasets.} We trained with 10\% of dataset for each domain. We show results on the validation sets.}
\vspace{-6mm} 
\end{table}
\subsection{In-Domain Annotation}
This section compares the performance of our model with state-of-the-art models and baseline models. We evaluate automatic and interactive modes using the Cityscapes dataset. We chose the Cityscapes dataset as it is one of the most comprehensive benchmarks available to evaluate instance segmentation performance. The Cityscapes dataset instance segmentation consists of eight classes. There are $2,975$, $500$, and $1,525$ for training/validation/test images. \vphantom{There are $2,975$ training images, $500$ validation images, and $1,525$ test images.} 
\\[7pt]
\noindent \textbf {Automatic Mode:} We perform two experiments in automatic mode. First, we compare the performances of our model and the state-of-the-art models on the Cityscapes test set. Here, we use Faster R-CNN \cite{ren2015faster} as an object detection model, which is one of the commonly used detection models. According to the results in Table \ref{tab_sota}, our model presents $29.6$ AP, which is very competitive compared to the state-of-the-art models using the pixel-wise approach. Our score shows the highest among the models, only trained with the Cityscapes dataset (\texttt{fine}), followed by SSAP \cite{gao2019ssap} and AdapIS \cite{sofiiuk2019adaptis}. Moreover, our model shows a $16 \%$ increase when compared to PolygonRNN++ \cite{acuna2018efficient}, a similar model applying the polygon-based approach. \linebreak
\indent Second, we compare our model with the baseline models using the Cityscapes validation set. The results of mIoU, F$_\mathrm{1px}$, and F$_\mathrm{2px}$ are reported in Table \ref{tab_auto}. We find that our model presents the highest metrics of $76.6$ mIoU, $52.5$ F at $1$ pixel, and $67.5$ F at $2$ pixels, which are even better than Spline-GCN, which has the highest performance among the baseline models. As shown in Figure \ref{fig:auto_pic_1} and \ref{fig:auto_pic_2}, Split-GCN exhibits satisfying results even when the object is disconnected.
\\[7pt]
\noindent \textbf {Points \& Split :} \label{points&split}
We evaluate how our model's accuracy varies according to control points $N$ and split factor $k$ on the Cityscapes validation set. The results of the first experiment about the split factor $k$ are in Table \ref{tab_split}. In Table \ref{tab_split}, $k=3$ shows the best performance. Interestingly, when the value of $k$ gets bigger than $3$, the performance decreases. We think that this is because the model concentrates more on locating the disconnected components rather than deforming the object boundary. The second experiment evaluates the performance according to the different number of control points $N$. As reported in Table \ref{tab_vertex}, we increase the control points by $10$ units, where the range of $N$ is $20 \leq N\leq 60$. From the results in Table \ref{tab_vertex}, we can see that the best performance is where $N$ is $40$.\\[7pt]
\noindent \textbf {Ablation Study :}
Using the Cityscapes validation set, we evaluate how our model's performance varies depending on our proposed motion vector branch, separating network, and interactive learning. Based on the experiment results on Table \ref{tab_split} and \ref{tab_vertex}, we set $k=3$ and $N=40$ as our best performing model. Table \ref{tab_abl} summarizes the results. The performance increases up to $8.2\%$ from the GCN as we apply the motion vector branch and separating network, and when interactive learning is applied, the performance increases an additional $2.2\%$.
\\[7pt]
\noindent \textbf {Interactive Mode:}
We test how our model performs when human interaction involves using the Cityscapes validation set (see Table \ref{tab_human}). In interactive mode experiments, the annotator continuously corrects the model's predicted vertices until meeting two independent conditions. The first condition is either reaching the performance of $1.0$ mIoU or reaching the maximum performance (see Figure \ref{fig:graph_1}). The second condition is whether locating the vertices more than five times to correct the segmentation (see Figure \ref{fig:graph_2}).\linebreak
\indent In Figure \ref{fig:graph_1}, our model reaches higher mIoU with fewer clicks than the two baseline models, PolygonRNN++ and Spline-GCN. This result implies that the annotator generates the ground-truth much quicker. In particular, our model improves about 1.1\% mIoU per click than Spline-GCN on average.\vphantom{Furthermore, our model takes three milliseconds on average in interactive inference.} We further compare the interactive mode with the automatic mode by each class to demonstrate human-in-the-loop performance in Figure \ref{fig:graph_2}. Consequently, our model's interactive mode improves mIoU up to $12.2$\% than the automatic mode with only five or fewer clicks.

\subsection{Cross-Domain Annotation}
In this section, we verify the generalization ability of our model. The generalization ability is crucial for the utilization of image annotation in various fields. We also demonstrate how quickly our model can be applied when fine-tuned with only $10\%$ of data from the new domain.
For evaluation, we use our model trained on the Cityscapes dataset following \cite{acuna2018efficient,ling2019fast}. We analyze our model's performance on four different data domains: KITTI \cite{geiger2012we} (urban street scenes), Rooftop \cite{sun2014free} (aerial images of a rural scene), ADE20K \cite{zhou2017scene} (general scene images), and Card.MR \cite{suinesiaputra2014collaborative} (medical images).
%
\\[7pt]
\noindent\textbf{Qualitative Results:} 
Table \ref{tab_cross} shows the results of comparing our model with the baseline models on the cross-domain datasets. In the KITTI and ADE, Split-GCN ($k=3$) outperforms when compared to $k=1$ or the other models. In the Rooftop dataset, Split-GCN ($k=1$) shows the performance at most $2.4\%$ higher than Spline-GCN, and in Card MR, $14.1\%$ higher than PolygonRNN++. Especially for these two datasets, Split-GCN with $k=3$ generates even better performance with $1.4\%$ higher than $k=1$. We demonstrate that Split-GCN adapts well to the new domains, even with limited training. 
\section{Conclusions}

In this paper, we proposed Split-GCN, which predicts points to be located uniformly at the object boundary. Our model locates the disconnected location and deforms the object boundary simultaneously to make sophisticated prediction, even when the object is disconnected. Moreover, we suggested the motion vector branch, which predicts the object boundary more precisely by informing destination information to the model. Our experiments confirmed the competitive performance with $29.6$ AP, $76.6$ mIoU, $52.5$ F$_\mathrm{1px}$, and $67.5$ F$_\mathrm{2px}$ scores on the Cityscapes dataset. Our model also demonstrates a compelling generalization ability by comprehending the contexts of new domain datasets. Future research should develop the technique to dynamically change the number of control points for a more detailed prediction of an object boundary. 

\clearpage

{\small
\bibliographystyle{ieee_fullname}
\bibliography{real_bib}
}

\end{document}